\documentclass[runningheads]{llncs}
\usepackage[T1]{fontenc}
\usepackage{graphicx}
\usepackage{booktabs}
\usepackage{multirow}
\usepackage{amsmath}
\usepackage{tabularx}
\usepackage{amssymb}
\usepackage{siunitx}

\newcolumntype{Y}{>{\centering\arraybackslash}X}
\begin{document}
\title{
Enabling Text-free Inference in \\
Language-guided Segmentation of Chest X-rays \\
via Self-guidance
}
\author{
    Shuchang Ye \and
    Mingyuan Meng \and
    Mingjian Li \and 
    Dagan Feng \and
    Jinman Kim
}

\institute{
    The University of Sydney
}
\maketitle              % typeset the header of the contribution
\begin{abstract}
Segmentation of infected areas in chest X-rays is pivotal for facilitating the accurate delineation of pulmonary structures and pathological anomalies. Recently, multi-modal language-guided image segmentation methods have emerged as a promising solution for chest X-rays where the clinical text reports, depicting the assessment of the images, are used as guidance. Nevertheless, existing language-guided methods require clinical reports alongside the images, and hence, they are not applicable for use in image segmentation in a decision support context, but rather limited to retrospective image analysis after clinical reporting has been completed. In this study, we propose a self-guided segmentation framework (SGSeg) \footnote{the code repository can be accessed at https://github.com/ShuchangYe-bib/SGSeg} that leverages language guidance for training (multi-modal) while enabling text-free inference (uni-modal), which is the first that enables text-free inference in language-guided segmentation. We exploit the critical location information of both pulmonary and pathological structures depicted in the text reports and introduce a novel localization-enhanced report generation (LERG) module to generate clinical reports for self-guidance. Our LERG integrates an object detector and a location-based attention aggregator, weakly-supervised by a location-aware pseudo-label extraction module. Extensive experiments on a well-benchmarked QaTa-COV19 dataset demonstrate that our SGSeg achieved superior performance than existing uni-modal segmentation methods and closely matched the state-of-the-art performance of multi-modal language-guided segmentation methods. 

\end{abstract}
\section{Introduction}

Chest X-rays play an essential role in the diagnosis of some pulmonary infectious diseases. In the analysis of chest X-rays, segmentation of the infected areas is essential for improving diagnostic accuracy, optimizing treatment plans, and enabling disease progression monitoring~\cite{segment_anything_medical}. However, manual segmentation conducted by radiologists is labor-intensive and prone to inconsistencies, posing challenges to its scalability and uniformity in clinical applications~\cite{seg_based_classification}. This inspires the integration of deep learning into the segmentation of chest X-rays, offering a pathway to automate the delineation process and enhance the efficiency and reliability of pulmonary diagnosis~\cite{seg_ct_cov19,seg_natural_medical}. The development of medical segmentation has been significantly advanced since the invention of U-Net~\cite{unet,review_unet}. With the progression of neural network architectures, U-Net has been extended as many variants (e.g., U-Net++~\cite{unet++}, Attention U-Net~\cite{att_unet}, Trans U-Net~\cite{trans_unet}, and Swin U-Net~\cite{swin_unet}) and obtained improved performance. Nevertheless, a persistent challenge within the medical domain remains: the inherent complexity of medical images poses difficulties for models to interpret underlying pathologies and identify disease locations, resulting in suboptimal segmentation accuracy for pulmonary lesions.

Recently, multi-modal learning has provided evidence that integrating visual and textual data exhibits superior performance over their uni-modal counterparts~\cite{multi>single}. Visual Language Pre-training (VLP)~\cite{vlp} has significantly advanced across various computer vision tasks by effectively bridging image and text features. For instance, CLIP~\cite{clip} adopted contrastive learning techniques to align the representations of image and text in latent space, fostering robust cross-modal similarities. Existing VLP primarily trained encoders, yet for tasks requiring both an encoder and a decoder, such as segmentation, a more comprehensive training approach to simultaneously optimize both components is essential. In medical image segmentation, LViT~\cite{lvit} demonstrated that the models guided by additional textual information can achieve higher performance. Building upon the LViT, Zhong et al.~\cite{languidemedseg} advanced image-text fusion techniques by introducing a text-guided decoder in U-Net. These multi-modal language-guided methods outperformed existing uni-modal methods, achieving state-of-the-art performance in chest X-ray segmentation. However, existing multi-modal methods necessitate the input of textual reports alongside images during inference, diverging from the clinical protocol of analyzing images prior to generating reports and thus reducing their clinical applications.

In this study, we explore leveraging linguistic context during training while enabling text-free inference in language-guided segmentation of chest X-rays. Our main contributions are summarized as follows:

\begin{itemize}
    \item We propose a self-guided segmentation framework (SGSeg) where the encoder-decoder process is self-guided by generated clinical reports during inference. 
    \item Our SGSeg introduces a novel Localization-Enhanced Report Generation (LERG) module that can accurately identify disease locations and generate reports to provide guidance for segmentation by utilizing object predictions from an object detector to assist the report generation process.
    \item To address the issue where most object predictions produced by object decoders are categorized as ``no class," with only a minor portion accurately indicating the locations of infected areas, we proposed a location-based attention aggregator to transform sparse object prediction into location features.
    \item To provide weak supervision of LERG, a clustering-based location-aware pseudo-label extractor is introduced to extract location information from clinical reports. 
\end{itemize}

Extensive experiments on the well-benchmarked QaTa-COV19 dataset demonstrate that the proposed SGSeg outperforms existing uni-modal inference segmentation methods and closely approximates the benchmarks set by the state-of-the-art multi-modal inference segmentation methods.

\section{Method}

\subsection{Self-guided segmentation framework (SGSeg)}

\begin{figure}
\centering
\includegraphics[width=0.95\textwidth]{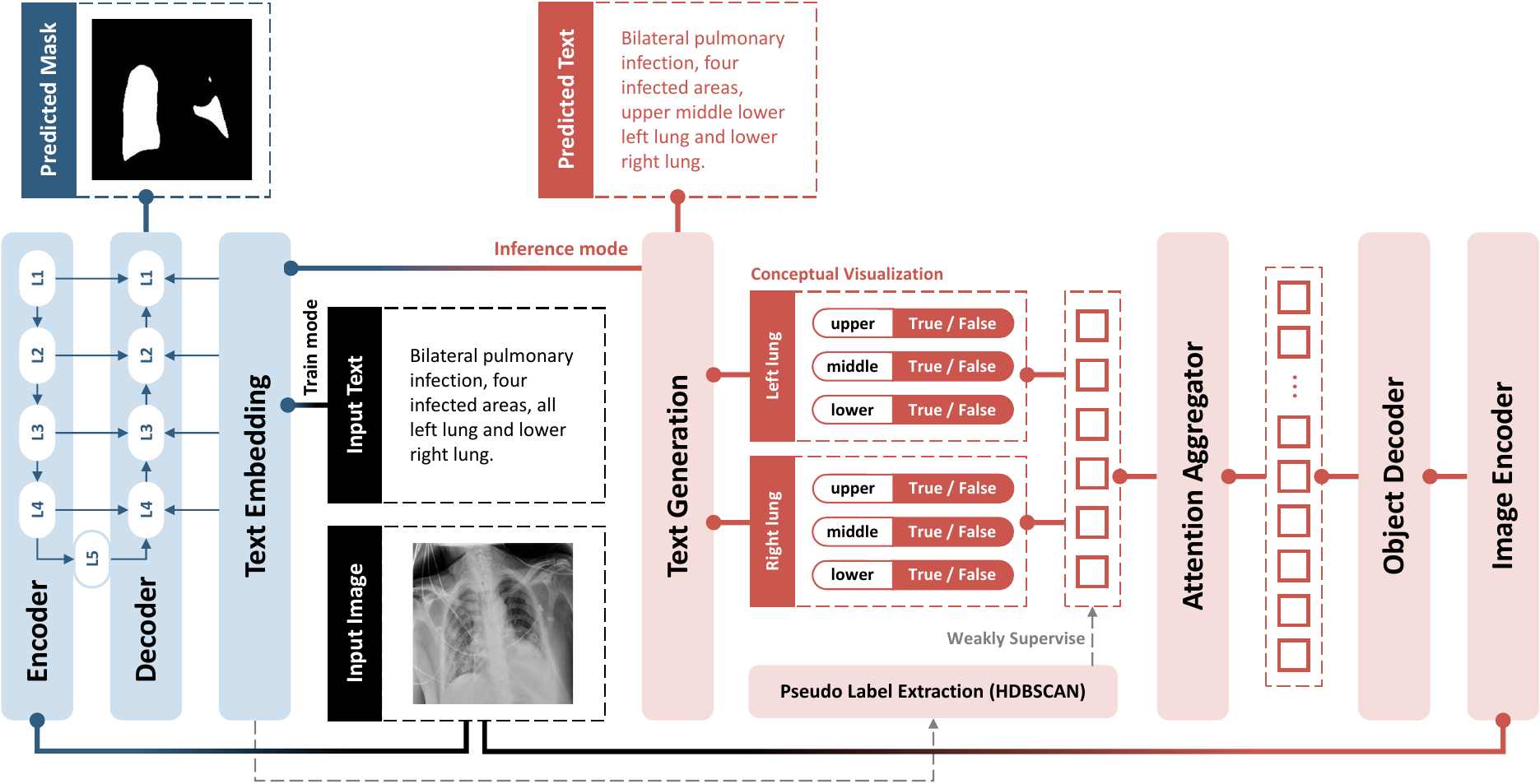}
\caption{The neural network architecture of the proposed SGSeg framework: The blue segment represents the Language-guided U-Net, while the pink segment denotes localization-enhanced report generation processes.} \label{architecture}
\end{figure}

The proposed Self-Guided Segmentation framework (SGSeg) comprises two main components: a Language-guided U-Net and a novel weakly-supervised localization-enhanced report generation (LERG) component (see Fig.~\ref{architecture}). During the training phase, ground-truth reports served as inputs to the text encoder from which labels were extracted to provide weak supervision for LERG. In the inference phase, generated reports replaced the ground truth as inputs to the text encoder, facilitating text-free inference.

\subsection{Language-guided U-Net}

In the Language-guided U-Net, the image downsampling process utilized Con-vNeXt-T~\cite{convnext}, pre-trained on ImageNet-1K, to sequentially reduce the dimensionality from 224x224 to 7x7 by a factor of 4 at each step. For the image up-sampling decoder, we adopted the GuidedDecoder structure from LanGuideMedSeg~\cite{languidemedseg}, utilizing a cross-modal attention module to fuse extracted position information with image features effectively. The text encoder was implemented using BERT~\cite{med_bert}, which was pre-trained through masked language modeling~\cite{bert} and multimodal contrastive learning~\cite{clip} on the MIMIC dataset.

\subsection{Localization-enhanced Report Generation}

\subsubsection{Understanding Text's Role in Language-guided Segmentation}

\begin{figure}
\centering
\includegraphics[width=0.95\textwidth]{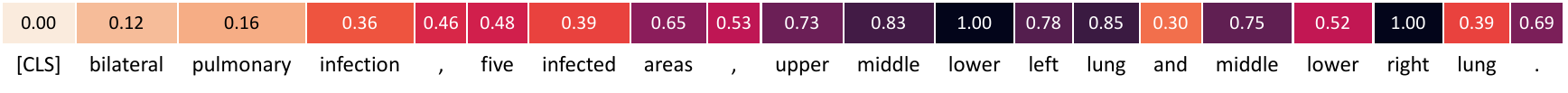}
\caption{Illustration of word importance, where we visualize the attention weight on each token of a report.} \label{q_times_k}
\end{figure}

To delve into the underlying principles of how text influences the segmentation results, we examined the importance of each word within a given text through a cross-modal attention module, as shown in Fig.~\ref{q_times_k}. The importance of each word was estimated based on the product of the query (q) and key (k) vectors for each word. The heatmap revealed a pronounced emphasis on location-descriptive words (``upper", ``middle", ``lower", ``left", and ``right") within the cross-modal attention framework. This phenomenon indicated the potential to refine segmentation accuracy by localizing infected areas.

\subsubsection{Location-aware pseudo-Label Extraction}

To effectively use the additional textual report during training, we first extract descriptions related to location information, utilizing a BERT model trained on X-ray to encode disease location into latent representation. Subsequently, we apply HDBSCAN~\cite{hdbscan} to group the text embedding into meaningful clusters that reflect the spatial relationships among reports.

\subsubsection{Weakly-supervised Localization-enhanced Report Generation}

We propose automatically generating reports focusing on spatial positioning to guide segmentation. Given the absence of ground truth labels for location prediction, pseudo-labels were extracted from reports to weakly-supervise the localization process. Our object detection network adhered to the RT-DETR~\cite{rtdetr} architecture, where images were first compressed via a CNN architecture, followed by intra-scale feature interaction through self-attention. Subsequently, features of varying granularity interact via a Cross-Scale Feature-Fusion Module (CCFM), with which object queries are decoded into object predictions via an object decoder, according to:

\begin{equation}
\label{query}
Q = Deocde(CCFM(F_{CNN}(I)))
\end{equation}

\noindent where $P$ represents object prediction derived from the image $I$ processed through ResNet50~\cite{resnet} backbone and the CCFM transformation. To refine the alignment between the predicted vector, denoted as $p$, and the pseudo-labels, represented by $y$, the Binary Cross-Entropy Loss was employed according to:

\begin{equation}
\label{loss}
Loss = -\frac{1}{N}\sum_{i=1}^N[y_i \cdot log(p_i)+(1-y_i) \cdot log(1-p_i)]
\end{equation}

\noindent where $N$ signifies the number of labels, specifically six for this framework, aligning with the ``upper", ``middle", and ``lower" regions across both lungs. Here, $p_i$ and $y_i$ denote the predicted probability and the actual label for the $i-th$ label, respectively.

The final step involves decoding the labels into precise infected area locations, enabling the inference of both the number of infected areas and the overall infection status across the lungs. This decoded information is then synthesized into a text description.

\subsubsection{Location-based Attention Aggregation}

This module is designed to refine sparse object predictions from the object decoder into location features by initializing a location-specific query vector $q$. The process involves calculating attention weights for each object prediction through matrix multiplication. Subsequently, we derive the aggregated features $A = softmax(Xq^T) \cdot X$ by the linear combination of these weighted object predictions, where $X$ denotes the input object predictions.

\section{Experiments}

\subsection{Dataset}

The dataset used to evaluate our methodology was the QaTa-COV19 dataset~\cite{qata} \footnote{The Qata-COVID-19 dataset used in this study can be accessed at the following URL: https://www.kaggle.com/datasets/aysendegerli/qatacov19-dataset.}, the only publicly available chest X-ray dataset with text medical reports. It comprises $9,258$ chest X-ray images of COVID-19 infections alongside segmentation annotations of corresponding infection regions. This dataset was augmented by Li et al. by providing brief, structured, textual descriptions detailing the infection site~\cite{lvit}. The dataset was partitioned adhering to the official split~\cite{languidemedseg} into $5,716$ for training, $1,429$ for validation, and $2,113$ for testing.

Subsequent refinement of the dataset \footnote{The refined dataset is available at https://github.com/ShuchangYe-bib/SGSeg/tree/main/data} involved the correction of erroneous descriptions, typographical errors, and ambiguous expressions, which impacted $4\%$ of the data. Pseudo-labels were derived from textual annotations to provide weak supervision of lesion localization. 

\subsection{Experiment Setup}

The performance of the proposed SGSeg was benchmarked against established uni- and multi-modal segmentation methods to demonstrate its effectiveness. Performance metrics for lesion localization and report generation were analyzed to understand the intermediary results and the mechanism enabling the replacement of textual input during inference. Furthermore, ablation studies were conducted to quantify the impact of individual components on the framework’s overall performance.

\subsection{Implementation Details}

We used the image size of $224 \times 224$ with a hidden dimension $768$ in the cross-attention module. PyTorch~\cite{pytorch} and PyTorch Lightning~\cite{pytorch_lightning} were used as the development environment with NVIDIA RTX A6000 GPUs. For training, we applied AdamW optimizer~\cite{adamw} with a cosine annealing learning rate policy (initial rate $3 \times 10^{-4}$) and reduced it to $< 1 \times 10^{-6}$. The batch size was set to $32$. Data augmentations, including random crops, masks, and rotations, were applied.

\section{Result and Discussion}

\subsection{Comparison with Existing Methods}

\begin{table}[t]
\centering
\caption{Performance comparison between our SGSeg and existing uni-modal and multi-modal segmentation methods on the QaTa-COV19 dataset. The best results of uni- and multi-modal methods are underlined. The results of our SGSeg are highlighted in bold.}
\label{metrics_comparison}
\begin{tabular}{llccc}
    \toprule
    Modality & Model & Accuracy & Dice & Jaccard \\
    \midrule
    \multirow{5}{*}{Uni-Modal} 
    & U-Net\cite{unet} & 0.945 & 0.819 & 0.692 \\
    & U-Net++\cite{unet++} & 0.947 & 0.823 & 0.706 \\
    & Attention U-Net\cite{att_unet} & 0.945 & 0.822 & 0.701 \\
    & Trans U-Net\cite{trans_unet} & 0.939 & 0.806 & 0.687 \\
    & Swin U-Net\cite{swin_unet} & \underline{0.950} & \underline{0.832} & \underline{0.724} \\
    \midrule
    \multirow{1}{*}{Multi-Modal Train,  Uni-Modal Inference} 
    & SGSeg (ours) & \textbf{0.971} & \textbf{0.874} & \textbf{0.778} \\
    \midrule
    \multirow{2}{*}{Multi-Modal}
    & LViT\cite{lvit} & 0.962 & 0.837 & 0.751 \\
    & LanGuideSeg\cite{languidemedseg} & \underline{0.975} & \underline{0.898} & \underline{0.815} \\
    \bottomrule    
\end{tabular}
\end{table}

\begin{figure}
\centering
\includegraphics[width=0.95\textwidth]{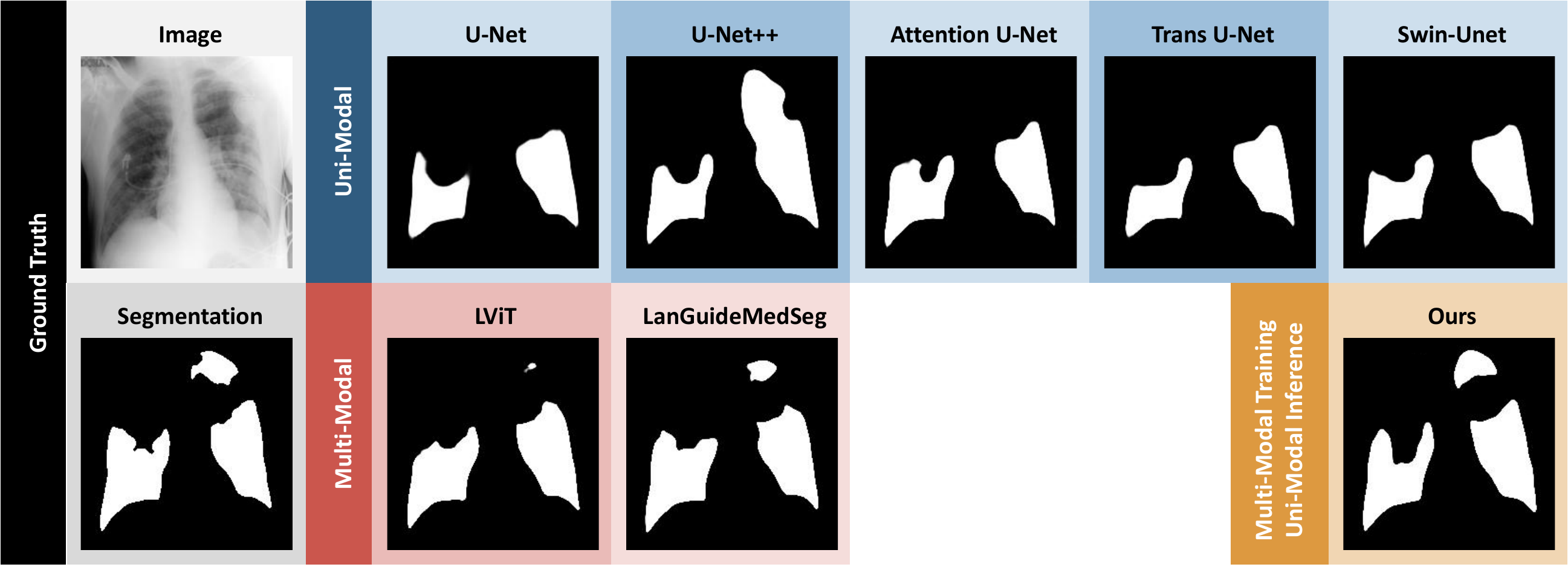}
\caption{Comparative Analysis of Segmentation Results: Uni-modal vs. Multi-modal Methods} \label{seg_comparison}
\end{figure}

To evaluate the SGSeg framework's efficacy, comparative analyses were conducted with current uni-modal and multi-modal segmentation models, as shown in table~\ref{metrics_comparison}. Results illustrate that SGSeg exceeds the performance of conventional uni-modal methods and closely matches that of advanced multi-modal approaches. Relative to the leading uni-modal inference model, our approach achieved a notable enhancement in accuracy from 0.950 to 0.971 ($2.21\%$), in the Dice coefficient from $0.832$ to $0.874$ ($5.05\%$), and in the Jaccard index from $0.724$ to $0.778$ ($7.46\%$). However, when compared with the top-performing multi-modal inference model, our method exhibited a slight decrease in accuracy by $0.41\%$, and reductions in the Dice coefficient and Jaccard index by $2.74\%$ and $4.75\%$, respectively. Fig.~\ref{seg_comparison} illustrated the significant impact of textual information on enhancing segmentation accuracy, particularly in challenging cases. It demonstrated that incorporating location-specific or pseudo-location data facilitates the model's precision in identifying pathological areas. The SGSeg model was trained to identify lesions through weak supervision provided by additional textual data, compensating for the absence of textual descriptions at inference by autonomously generating this essential information. Consequently, the model leveraged guidance from the detector to achieve enhanced segmentation outcomes.

\subsection{Ablation Study}

\begin{table}
\centering
\caption{Ablation studies on the impact of individual components in SGSeg: Without text - uni-modal segmentation; Visual-language Pre-training - pre-trained with CLIP and fine-tuned by uni-modal segmentation; Self-Guidance - using generated text as input during inference; Full Text - using ground truth text as input during inference.}
\label{ablation_study}
\begin{tabular}{lccc}
    \toprule
     & Accuracy & Dice & Jaccard \\
    \midrule
    Without Text & 0.953 & 0.846 & 0.725 \\
    Vision-language pre-training (CLIP) & 0.962	& 0.850	& 0.740 \\
    Self-Guidance (simple report generation) & 0.966 & 0.861 & 0.759 \\
    Self-Guidance (weakly-supervised LERG) & 0.971 & 0.874 & 0.778 \\
    Full Text & 0.973 & 0.890 & 0.797 \\
    \bottomrule    
\end{tabular}
\end{table}

Our framework leverages additional language information during training while eliminating the need for textual input during inference. To demonstrate the utility of synthetic text and validate our self-guided design, we conducted an ablation study (Table~\ref{ablation_study}). The results show that including additional textual information significantly improves segmentation performance. Comparing vision-language pre-training with our method, our multi-modal encoder-decoder training outperformed multi-modal encoder-focused pre-training. SGSeg significantly outperformed methods without textual input and closely matched those using ground truth text for inference. This indicates that our self-guidance framework effectively utilizes text for weak supervision during training and autonomously generates text inputs for inference. Additionally, the proposed weakly supervised LERG module enhanced segmentation accuracy, underscoring the efficacy of incorporating a location-aware pseudo-label extractor and a location-based attention aggregator.

\begin{figure}
\centering
\includegraphics[width=0.95\textwidth]{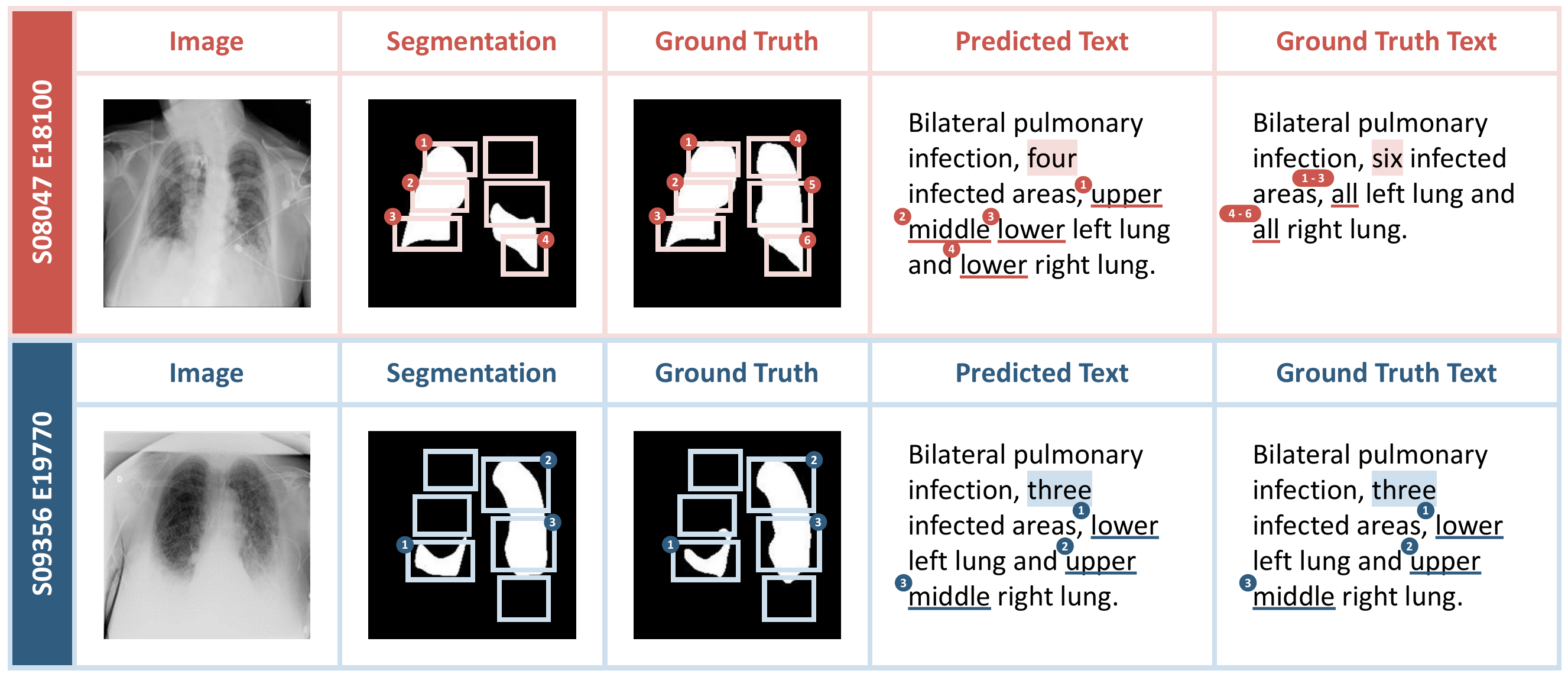}
\caption{Comparative analysis on the relation between generated text and segmentation outcomes} \label{interpretability}
\end{figure}

\begin{figure}
\centering
\includegraphics[width=0.95\textwidth]{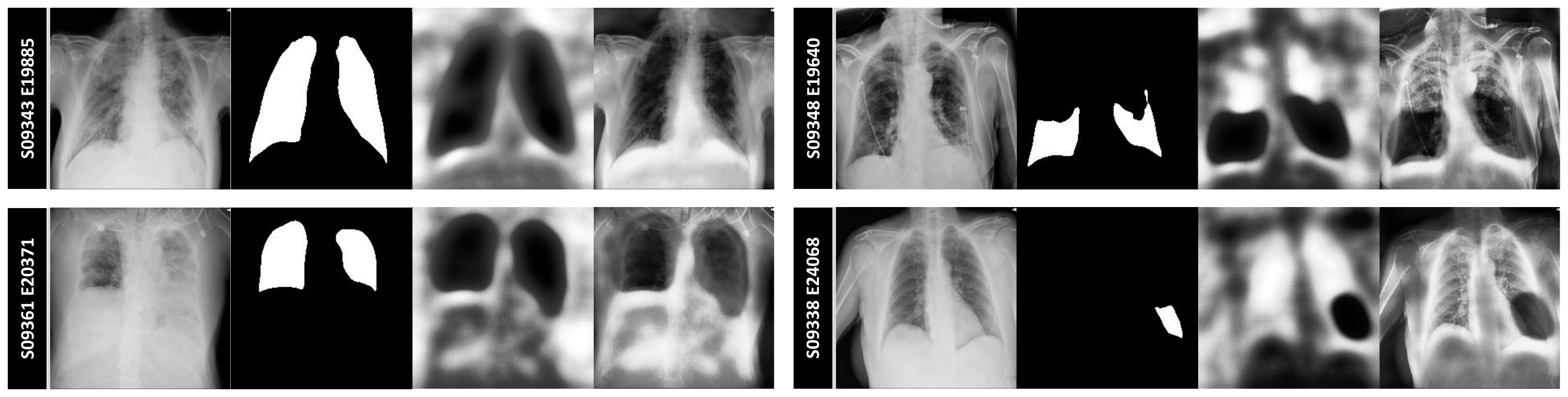}
\caption{Visualization of the model's attention distribution over the input image arranged sequentially as follows: image, ground truth segmentation, attention map, and attention map projected onto the image} \label{attention_map}
\end{figure}

\subsection{Visualization}

We conducted an in-depth examination of the attention maps and the relationship between the generated text and the segmentation outcomes. Our model demonstrates accurate attention to lesions, which is fundamental to segmentation (see Fig.~\ref{attention_map}). Notably, the transition from exposure to ground truth text during training to reliance on generated text during inference impacts segmentation outcomes. Comparative analysis, as depicted in Fig.~\ref{interpretability}, indicates that inaccuracies in generated reports moderately influence the model’s performance.

\section{Conclusion}

This study identified a crucial shortfall in current language-guided segmentation methods: their reliance on textual inputs during inference diminishes their relevance in clinical practice. To overcome this challenge, we analyzed the text’s role in language-guided segmentation and proposed an innovative self-guided segmentation framework tailored for text-free analysis. Experiments on the QaTa-COV19 dataset showed that our SGSeg significantly outperformed existing uni-modal image-only methods and closely approached the multi-modal methods requiring text reports during inference.


\end{document}